\documentclass[10pt,twocolumn,letterpaper]{article}

\usepackage{iccv}              
\usepackage{multirow}
\usepackage{algorithm}
\usepackage{algpseudocode}
\usepackage{makecell}
\definecolor{iccvblue}{rgb}{0.21,0.49,0.74}
\usepackage[pagebackref,breaklinks,colorlinks,allcolors=iccvblue]{hyperref}

\def\paperID{11935} 
\def\confName{ICCV}
\def\confYear{2025}

\title{I2V-GS: Infrastructure-to-Vehicle View Transformation with Gaussian Splatting for Autonomous Driving Data Generation}

\author{%
  Jialei Chen$^{1}$ \quad
  Wuhao Xu$^{2}$ \quad
  Sipeng He$^{3}$ \quad
  Baoru Huang$^{4}$ \quad
  Dongchun Ren$^{1}$\\[0.5em]
  {\normalsize%
    $^{1}$Yootta \qquad
    $^{2}$Soochow University \qquad
    $^{3}$Southeast University \qquad
    $^{4}$University of Liverpool}%
}

\begin{document}
\maketitle

\begin{abstract}
Vast and high-quality data are essential for end-to-end autonomous driving systems. However, current driving data is mainly collected by vehicles, which is expensive and inefficient. A potential solution lies in synthesizing data from real-world images. Recent advancements in 3D reconstruction demonstrate photorealistic novel view synthesis, highlighting the potential of generating driving data from images captured on the road. This paper introduces a novel method, I2V-GS, to transfer the \textbf{I}nfrastructure view \textbf{T}o the \textbf{V}ehicle view with \textbf{G}aussian \textbf{S}platting. Reconstruction from sparse infrastructure viewpoints and rendering under large view transformations is a challenging problem. We adopt the adaptive depth warp to generate dense training views. To further expand the range of views, we employ a cascade strategy to inpaint warped images, which also ensures inpainting content is consistent across views. To further ensure the reliability of the diffusion model, we utilize the cross-view information to perform a confidence-guided optimization. Moreover, we introduce RoadSight, a multi-modality, multi-view dataset from real scenarios in infrastructure views. To our knowledge, I2V-GS is the first framework to generate autonomous driving datasets with infrastructure-vehicle view transformation. Experimental results demonstrate that I2V-GS significantly improves synthesis quality under vehicle view, outperforming StreetGaussian in NTA-Iou, NTL-Iou, and FID by 45.7\%, 34.2\%, and 14.9\%, respectively.


\end{abstract}
\begin{figure*}[!h]
    \centering
    \includegraphics[width=\linewidth]{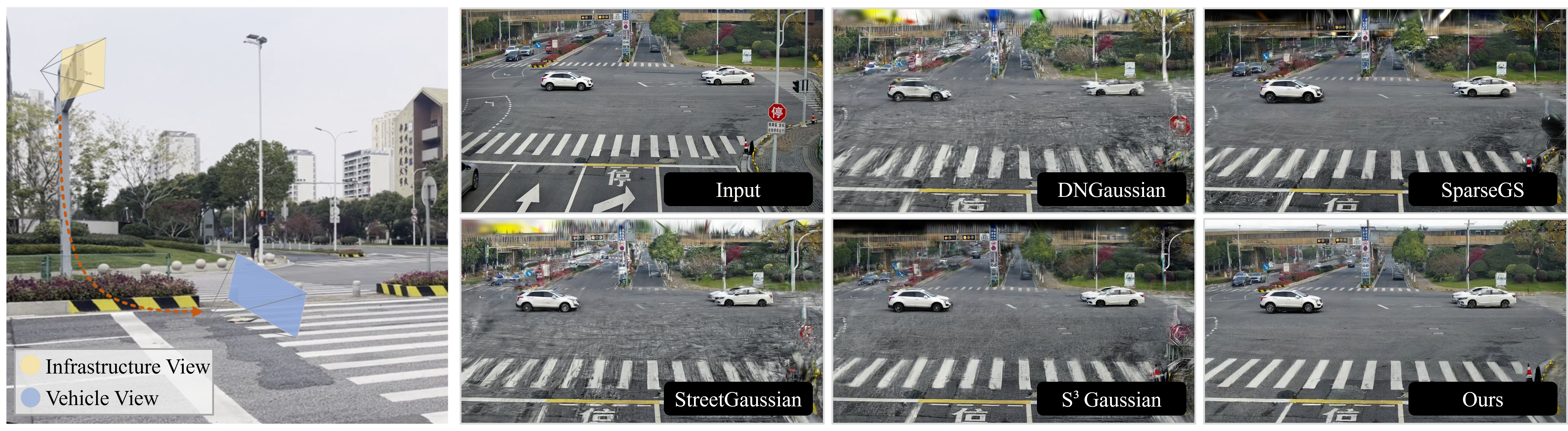}
    \caption{We propose transferring images captured by infrastructure to vehicle views for autonomous driving system training to reduce the cost of data collection \textbf{(left)}. Previous Gaussian Splatting methods face challenges in synthesizing vehicle views. In contrast, our approach significantly improves the image quality \textbf{(right)}. }
    \label{fig:overview}
\end{figure*}

\section{Introduction}

In recent years, there has been great progress in end-to-end autonomous driving systems \cite{hu2022st,hu2023planning}, which convert sensor inputs to control signals directly. However, one of the major challenges for end-to-end autonomous driving is the need for vast training data to achieve reliable performance \cite{chen2024endtoendautonomousdrivingchallenges}. These large datasets are essential for training models capable of handling complex and dynamic environments.

Currently, the primary methods for acquiring autonomous driving data can be divided into dedicated data collection fleets, production vehicle fleets, and synthetic datasets \cite{chen2024endtoendautonomousdrivingchallenges}. Nevertheless, these methods face significant challenges. Data collection via dedicated fleets, such as the Waymo \cite{sun2020scalability} and nuScenes \cite{caesar2020nuscenes} datasets, provides realistic environmental data, but it is relatively expensive due to high operational costs associated with the vehicles, sensors, and safety drivers. On the other hand, production vehicle fleets, like Tesla's, generate vast amounts of real-world driving data but face problems related to data privacy and high data transmission costs \cite{DBLP:journals/corr/abs-2209-04022}. These limitations have led to the increasing adoption of dataset synthesis techniques, which offer a cost-effective and efficient alternative while capable of producing diverse scenarios. Synthetic datasets typically can be divided into utilizing game engines for rendering data \cite{ros2016synthia,li2021learning}, generating driving scenarios with generative models \cite{hu2023gaia, gao2023magicdrive, wang2024driving, yang2024generalized, wang2023drivedreamer, gao2024vista, swerdlow2024street}, and viewpoint transformations or adding additional objects based on real-world dataset \cite{zhou2024drivinggaussian,huang2024textit,yan2024street,yang2024driving,zhao2024drivedreamer4d}. However, transitioning from synthetic data to real-world applications presents significant challenges, as algorithms must effectively bridge the domain gap to ensure their performance remains robust in real-world environments \cite{liu2024survey}. Synthesis datasets from images captured in real-world~\cite{yang2024driving,zhao2024drivedreamer4d} are constrained by the size of the original collected data, limiting their scalability.

Considering the efficiency and validity, collecting information from infrastructure sensors and then transforming into vehicle views for autonomous driving system training is more effective and reliable, as illustrated in Fig. \ref{fig:overview}. Given roadside cameras capture images continuously, this approach enables the synthesis of a virtually unlimited number of datasets, improving the efficiency of data collection.


Nevertheless, in infrastructure-to-vehicle (I2V) view transformation tasks, sparse viewpoints and large view transformation pose challenges in rendering novel vehicle views. Previous autonomous scene reconstruction methods \cite{zhou2024drivinggaussian,huang2024textit,yan2024street} leverage the motion of vehicles to obtain multi-view images. However, this approach is not applicable in the sparse and fixed viewpoints setting. Sparse view reconstruction methods \cite{xiong2023sparsegs,li2024dngaussian} introduce depth prior as a constraint. However, they fail to render high-quality images under large view transformation.

To address these challenges, we propose I2V-GS. Our approach first calibrates monocular depth with LiDAR and employs the adaptive depth warp to provide a dense training view. To further expand the range of views, we utilize the diffusion model to inpaint holes in warped images, where a cascade strategy is adopted to ensure the consistency of inpainting content across views. To further improve the reliability of the diffusion model, we utilize cross-view information to assess the inpaint content to carry out a confidence-guided optimization for pseudo views. Moreover, we introduce RoadSight, a multi-modality, multi-view dataset from real scenarios for I2V view transformation. To our knowledge, I2V-GS is the first framework to generate autonomous driving datasets with I2V view transformation. As shown in Fig.~\ref{fig:overview}, our approach enhances the novel view synthesis quality in vehicle view, achieving a relative improvement in NTA-Iou, NTL-Iou, and FID by 45.7\%, 34.2\%, and 14.9\%, respectively, comparing with StreetGaussian~\cite{yan2024street}.


The main contributions of this work are as follows: 
\begin{itemize}
    \item We present I2V-GS, the first framework that generates autonomous driving datasets with infrastructure-vehicle view transformation. 
    \item We propose the adaptive depth warp to generate dense training views, enabling rendering high-quality images under sparse view input and large view transformation settings. 
    \item We introduce the cascade diffusion strategy to guarantee content consistency among pseudo views and leverage cross-view information in confidence-guided optimization for reliable inpaint content.
\end{itemize}

\section{Related Work}
\subsection{Novel View Synthesis from Sparse View}

Recent advances in 3D Gaussian Splatting (3DGS) have sought to address the sparse-view reconstruction challenge through two primary paradigms: depth-prior supervision and diffusion-based refinement. Depth-guided methods leverage geometric priors to compensate for insufficient multi-view constraints. Works like DNGaussian \cite{li2024dngaussian} employ monocular depth estimators \cite{yang2024depth} to constrain Gaussian positions. While these approaches mitigate floaters and improve surface coherence, their reliance on scale-ambiguous monocular predictions can lead Gaussians to distribute to suboptimal positions. Diffusion-based methods leverage generative models to predict missing details. Deceptive-NeRF \cite{liu2023deceptive} pioneers this direction by iteratively refining neural radiance fields using diffusion-model to refine rendered novel views. Subsequent 3DGS adaptations like SparseGS \cite{xiong2023sparsegs} apply score distillation sampling (SDS) \cite{poole2022dreamfusion} to align Gaussian renderings with diffusion priors. While effective for detail synthesis, these methods suffer from content inconsistency in unseen areas.

\subsection{Driving Scene Synthesis}
\textbf{Reconstruction-based Method.} Early methods \cite{Tancik_2022_CVPR,turki2023suds} utilize neural radiance fields (NeRF) \cite{nerf} to reconstruct the driving scene. Though these methods achieve high-quality rendering results, they suffer from long training and inference time. Recently, 3DGS \cite{kerbl3Dgaussians} introduces an efficient process, that represents scenes with a set of anisotropic Gaussians and achieves high-quality rendering from sparse point cloud inputs with adaptive density control. Several works \cite{zhou2024drivinggaussian,huang2024textit,yan2024street,yan2024street} extend 3DGS to model driving scenes by decomposing the static background and dynamic objects. However, These methods can only render interpolate views, where sensor data closely matches the training data distribution, which is inadequate for training autonomous driving models. More recently, some works \cite{yang2024driving,zhao2024drivedreamer4d, ni2024recondreamer,yan2024streetcrafter} propose to adapt the diffusion model to reconstruct driving scenes and generate extrapolated views. However, these methods rely on the motion of vehicles to obtain multi-view images, limiting their applicability in I2V transformation tasks where viewpoints are fixed.

\textbf{Generative-based Method.} Recently, generative models have shown significant potential in generating unseen and future views based on the current frame. Many works \cite{hu2023gaia, gao2023magicdrive, wang2024driving, yang2024generalized, wang2023drivedreamer, gao2024vista, swerdlow2024street} extent video diffusion models into autonomous driving to simulate different driving scenarios. Though generating many scenarios, they fail to capture the underlying 3D model, leading to inconsistent geometry and texture in the generated videos.

\begin{figure*}[t]
    \centering
    \includegraphics[width=\linewidth]{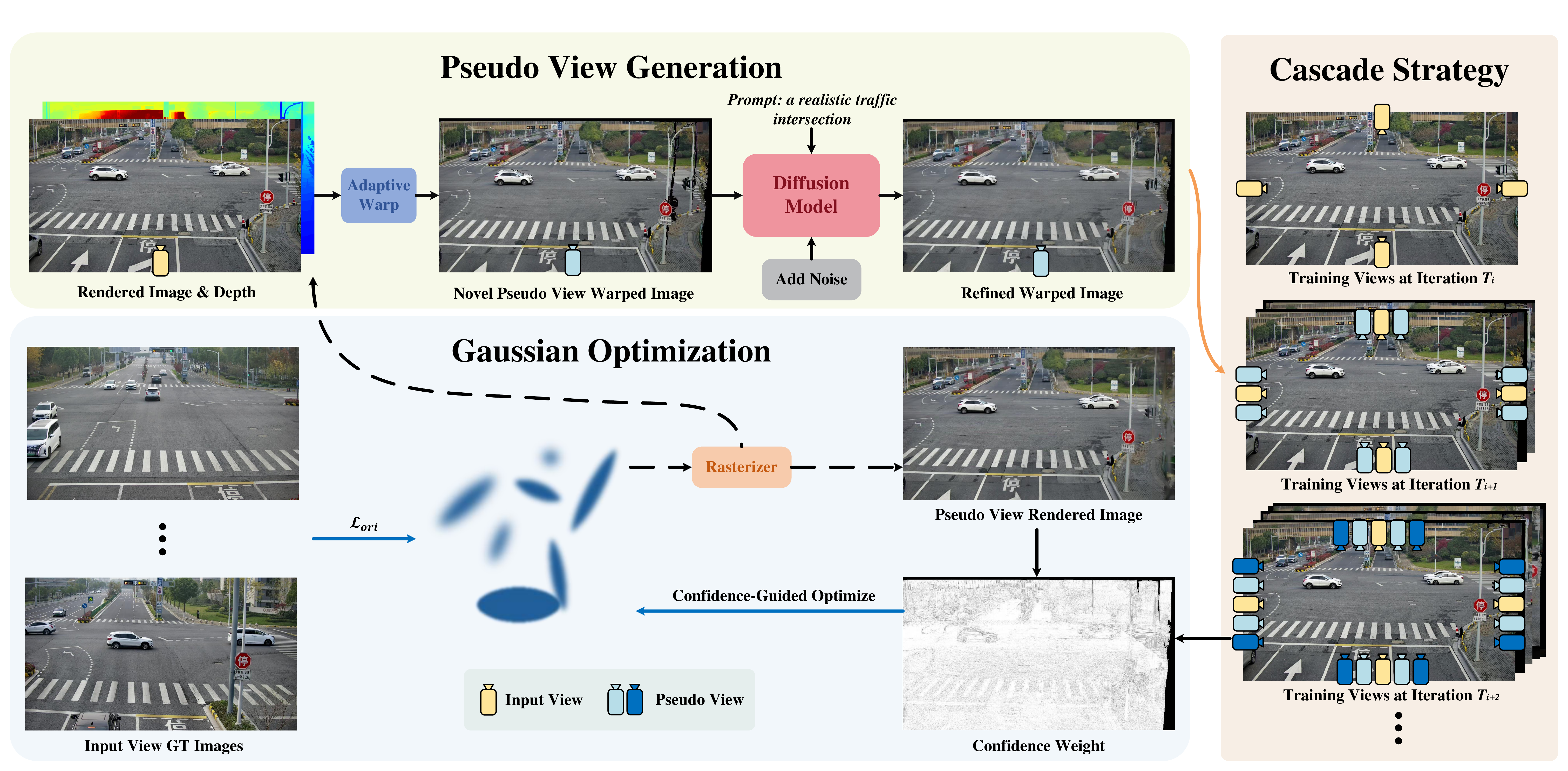}
    \caption{The overall framework of I2V-GS. Initially, the input views warm up the Gaussian optimization. Then, given the rendered image and depth of the current training view, the adaptive depth warp is employed to generate a novel pseudo view warped image, and the diffusion model is leveraged to inpaint unseen areas. Subsequently, the refined warped image would be added to training views and leverage the cross-view information to perform a confidence-guide optimization. Furthermore, we employ a cascade strategy to generate pseudo views progressively to ensure content consistency.}
    \label{fig:pipeline}
\end{figure*}

\section{Preliminary: 3D Gaussian Splatting}
3DGS~\cite{kerbl3Dgaussians} represents scenes with a set of differentiable 3D Gaussians. Each 3D Gaussians consists of learnable attributes: position $\mu$, rotation $r$, scaling $s$, opacity $o$, and spherical harmonic (SH) coefficients. Formally, the impact of a 3D Gaussian on location $x$ is defined by the Gaussian distribution:   
\begin{equation}
    G(x) = \exp\left( -\frac{1}{2} (x - \mathbf{\mu})^\mathrm{\textit{T}} \Sigma^{-1} (x - \mathbf{\mu}) \right),
    \label{eq:gaussian_pdf}
\end{equation}
where $\Sigma$ is the covariance matrix, which can be decomposed into $ \Sigma = R S S^\mathrm{\textit{T}}  R^\mathrm{\textit{T}}$, $R \in \mathbb{R}^4$ is a rotation matrix expressed with quaternions, and $S \in \mathbb{R}^3$ is a diagonal scaling matrix. The 3D Gaussian is projected onto the 2D image planes for rendering, where the projected 2D Gaussian is sorted by its depth value. The final rendering equation for the color $\hat{C}(X)$ of each pixel $X$ is: 
\begin{equation}
    \hat{C}(X) = \sum_{i \in N} c_i \alpha_i \prod_{j=1}^{i-1} (1 - \alpha_i),
    \label{eq:placeholder}
\end{equation}
\begin{equation}
    \alpha_i =  o_iG_i^{2D}(p),
\end{equation}
where $c_i$ is the color defined from the SH coefficients and $\alpha_i$ is the density calculated by multiplying the projected 3D Gaussian with the opacity $o_i$. The covariance matrix after projection is calculated by $\Sigma' = J W \Sigma W^T J^T$, where $J$ is the Jacobian of the affine approximation of the projective transformation and $W$ represents the viewing transformation matrix.

The 3D Gaussians $\mathcal{G}$ is optimized by the combination of RGB loss, depth loss, and SSIM loss:
\begin{equation}
\mathcal{L}_{ori} = \lambda \cdot \| \hat{I} - I \|_1 + (1-\lambda)  \cdot \text{SSIM}(\hat{I}, I)+  \| \hat{D} - D \|_1 ,
\label{eq:original_optimization}
\end{equation}
where $\hat{I}$ and $I$ refer to rendered and ground truth image, $\hat{D}$ and $D$ represent rendered and ground truth depth, SSIM($\cdot$) is the operation of the Structural Similarity Index Measure, and $\lambda$ is the loss weight.

\section{I2V-GS}
As shown in Fig. \ref{fig:dataset}, in the I2V view transformation task, sparse viewpoints and large view transformation cause the rendering of novel vehicle views difficult. To address these challenges, we propose a novel framework, I2V-GS. As is shown in Fig.~\ref{fig:pipeline}, we first warm up the Gaussian optimization with sparse input views. Then, we present a cascade pseudo view generation method to provide dense training views. Specifically, we utilize LiDAR to calibrate monocular depth to provide a real depth and propose the adaptive depth warp to generate proper pseudo views (Sec. \ref{section:Adaptive Depth Warp} and \ref{section:Mono-depth with LiDAR Prior}). The cascade strategy is employed to inpaint holes in warped images iteratively to guarantee content consistency and enable a wide range of training views (Sec. \ref{section:Cascade Diffusion Strategy}). To ensure the inpaint content aligns with the real world, the cross-view information is applied to assess the inpaint content in optimization (Sec. \ref{section:Confidence-Guided Optimisation}).
\subsection{Adaptive Depth Warp} 
\label{section:Adaptive Depth Warp}
To provide dense training views, we generate pseudo views $\mathcal{V}'$ around the input views $\mathcal{V}$ via forward warping $\psi$. Specifically, we project rendered 3D points $p=(x,y,z)^T$ under viewpoint $\mathcal{V}_i$ to a novel view through:
\begin{equation}
p' = KR'R^{-1}K^{-1}p+K(T'-R'R^{-1}T)
\label{eq:forward_Warp}
\end{equation}
where $p$ is from the depth map obtain in Sec.~\ref{section:Mono-depth with LiDAR Prior}, $p'$, $R'$, and $T'$ are the target view's 3D points, rotation matrix, and translation vector, respectively. $K$ is camera intrinsic, $T$ is the source view's translation, and $R$ is the source view's rotation. Then, $p'$ can be projected onto the image plane  with:
\begin{equation}
\frac{1}{||w_{norm}||}
\begin{pmatrix}
u \\
v\\
1
\end{pmatrix}
= K
\begin{pmatrix}
R & T \\
0 & 1
\end{pmatrix}
\begin{pmatrix}
x \\
y\\
z \\
1
\end{pmatrix}
\label{eq:projection}
\end{equation}
where $||w_{norm}||$ is used to normalize the homogeneous coordinate and $(u,v)$ is the pixel coordinate. Then, $(u,v)$ are rasterized into $I_{warp}$, $D_{warp}$, and $M_{warp}$ via z-buffering and bilinear sampling.

Directly applying depth warping with fixed displacement and rotations often leads to two critical failure cases: 1) over-warping from excessive displacements that amplify geometric errors, causing distorted artifacts, and 2) under-warping from insufficient displacements that limit viewpoint variation. To balance this trade-off, we introduce an adaptive depth warp strategy constrained by pixel-level spatial consistency. For 3D points observed in the source view $\mathcal{V}$, let $(\Delta u,\Delta v)^\top$ denote the pixel displacement in the pseudo-view $\mathcal{V}'$. Replacing $(x,y,z,1)^\top$ in Eq. \ref{eq:projection} with $p$ and $p'$, we express the displacement relationship as follows:

\begin{equation}
\begin{pmatrix}
\Delta u \\
\Delta v
\end{pmatrix}
=
\frac{1}{z(z + \Delta t_z)}
\begin{pmatrix}
f_x \Delta t_x z + c_x \Delta t_z z - \Delta t_z x \\
f_y \Delta t_y z + c_y \Delta t_z z - \Delta t_z y
\end{pmatrix}
\label{eq:warp_pixel_difference}
\end{equation}
where $\Delta t=(\Delta t_x, \Delta t_y, \Delta t_z)^\top =T'-T$ represents the relative translation between views, and $f_x$, $f_y$, $c_x$ , $c_y$ are camera intrinsics. Given a pre-defined warp difference $\varepsilon$, we can formulate the equation as:
\begin{equation}
\left\| \begin{pmatrix} \Delta u \\ \Delta v \end{pmatrix} \right\|_{\infty} \leq \varepsilon
\label{eq:control_warp}
\end{equation}


Solving Eq.~\ref{eq:control_warp} involves parameters $\Delta t_x$, $\Delta t_y$, $\Delta t_z$, $z$, and $\varepsilon$. For practical optimization, we first set $\Delta t_z = 0$ to decouple horizontal/vertical shifts, which will also decouple $x$ and $y$ in Eq. \ref{eq:warp_pixel_difference}. After that, $z$ is set to the minimum scene depth $z_{min}$ as a conservative estimate. This yields simplified bounds as below, and the depth warp can be controlled by giving a proper $\varepsilon$ (see appendix for more details):

\begin{equation}
\left\{
\begin{aligned}
\Delta t_x &\leq \frac{\varepsilon z_{min}}{f_x} \\
\Delta t_y &\leq \frac{\varepsilon z_{min}}{f_y}
\end{aligned}
\right.
\label{eq:warp_result}
\end{equation}


\subsection{LiDAR-Anchored Monocular Depth Calibration}
~\label{section:Mono-depth with LiDAR Prior}
The performance of depth warp critically depends on the accuracy of scene geometry estimation. Nevertheless, 3DGS \cite{kerbl3Dgaussians} tends to generate imprecise geometries due to insufficient multi-view constraints in sparse view settings. To tackle this problem, we propose the monocular depth with LiDAR prior, which aligns monocular predictions with accurate LiDAR measurements, to provide precise geometry guidance.


Standard monocular depth estimation converts disparity $d_{mono}$ to depth  $D_{mono}$ via:
\begin{equation}
   D_{mono}=\frac{b_{train}\cdot f_{train}}{d_{mono}}
   \label{eq:disparity}
\end{equation}
where $b_{train}$ and $f_{train}$ are domain-specific parameters tied to the training data distribution, e.g. forward-facing vehicle cameras in Waymo \cite{sun2020scalability}.

However, infrastructure images differ significantly from the training data distribution, which may cause disparity prediction bias and depth offsets when applying Eq. \ref{eq:disparity}.
To address this misalignment, we propose a generalized formulation:
\begin{equation}
D_{align}=\frac{c_1}{d_{mono}+c_2}+c_3
    \label{eq:depth_align}
\end{equation}
where $c_1$ corresponds to the product $b\cdot f$ in Eq. \ref{eq:disparity}, ensuring consistency with the classical framework. $c_2$ and $c_3$ are used to rectify potential biases and offsets in the disparity and depth values.

Then, we leverage the LiDAR depth $D_{lidar}$ to conduct a pair sample $\mathcal{P} =\left\{ \left( D_{lidar}^{(i,j)}, d_{mono}^{(i,j)} \right) \mid D_{lidar}^{(i,j)} \neq 0 \right\}$ to optimize parameter $c_1$, $c_2$ and $c_3$, where $(i,j)$ denotes pixel coordinates. The optimization is processed via nonlinear least squares with Huber loss $\mathcal{L}_H$:
\begin{equation}
\min_{c_1, c_2, c_3} \sum_{\mathcal{P}} \mathcal{L}_{H} \left( D_{lidar} - \frac{c_1}{d_{mono}+c_2}-c_3 \right)
\end{equation}

\begin{algorithm}[t]
\caption{Cascade Strategy}
\label{alg:cascade_diffusion_strategy}
\begin{algorithmic}
\Require Initial Gaussian model $\mathcal{G}_0$, diffusion model $\mathcal{D}$, rasterization $R$, training iterations $T$, depth warp steps $T_w$
\For{$t = 0, ...,T-1$}
    \If{$t$ in $T_w$}
        \State $\hat I', \hat D' \gets \{R(\mathcal{G}_t, \mathcal{V}_{j-1}')\}_{j=1}^{F'}$
        \State $I_{warp}, D_{warp}, M_{warp} \gets \psi(\hat I', \hat D' \mid \mathcal{V}_{j-1}',\mathcal{V}_{j}' ) $
        \State $I’, D’, M’ \gets \mathcal{D}(I_{warp}, D_{warp}, M_{warp})$
        \State $\mathcal{V}_{j}'(gt) \gets I’, D', M’$
    \EndIf
    \State $\hat I_t, \hat D_t \gets \{R(\mathcal{G}_t, \mathcal{V}_{i})\}_{i=0}^F$
    \State Compute loss $\mathcal{L}$ $\gets$ $\mathcal{L}_{ori}$, 
    \State Backpropagate loss and update $\mathcal{G}_{t+1}$
    \State $\hat I_{t}' \gets \{R(\mathcal{G}_t, \{R(\mathcal{G}_t, \mathcal{V}_{j}')\}_{j=1}^{F'}$
    \State Compute loss $\mathcal{L}'$ $\gets$ $\mathcal{L}_{con}(\hat I_{t}',I_{gt}')$, 
    \State Backpropagate loss and update $\mathcal{G}_{t+1}$
\EndFor
\State \Return $\mathcal{G}_T$
\end{algorithmic}
\end{algorithm}

\subsection{Cascade Strategy}
\label{section:Cascade Diffusion Strategy}
One key challenge with depth warp is that it usually introduces occlusion holes in pseudo views. While diffusion models can inpaint missing regions, their stochastic nature causes inconsistent content across frames. In this paper, we propose a cascade strategy. As illustrated in Alg. \ref{alg:cascade_diffusion_strategy}, we first warm up the model with input sparse views. Then, we carry out depth warp to provide dense training views. Specifically, we generate pseudo views in a cascade manner. In each round generation, the novel pseudo view $\mathcal{V}_j$ is based on the prior pseudo view $\mathcal{V}_{j-1}$. Following the warp, a latent diffusion model \cite{rombach2022high} is employed to inpaint the occluded regions. In this process, both the warped images and their corresponding hole masks are first encoded into a latent space $h$, where the inpainting operation is performed to yield a refined latent representation $\hat{h}$. The final, refined pseudo view is obtained by decoding $\hat{h}$ to an RGB image. This cascade mechanism enables the propagation of information from prior views, ensuring the consistency of inpaint contents across views. Moreover, the conventional depth warp is inherently limited by the accumulation of occlusion holes that arise from large viewpoint shifts. In contrast, incorporating inpainting methods effectively mitigates these occlusions, thereby broadening the operational range of depth warping and enabling a wider range of training views.

\subsection{Confidence-Guided Optimization}
\label{section:Confidence-Guided Optimisation}
While diffusion-based inpainting helps complete occluded regions, the stochastic generation process may introduce semantic inconsistencies between pseudo-views and actual scenes. To mitigate this, we propose a confidence-guided optimization scheme that leverages multi-view consensus to weight supervision signals. We adopt the $L_2$ difference to detect pixel alignment and SSIM to evaluate perceptual similarity at the patch-structure level. Given the inpainted image $I'$ and rendered image $\hat I'$ under $\mathcal{V}_j'$, the combined confidence weight is:
\begin{equation}
    W = \lambda_1 \cdot(1- L_2(\hat I',I')) 
    \\
    + (1-\lambda_1) \cdot SSIM(\hat I',I')
\end{equation}
where $W\in(0,1)$ represents the confidence weight, with values closer to 1 indicating higher confidence. Then, the confidence-guided loss can be expressed as:

\begin{equation}
\mathcal{L}_{\text{con}} = \mathbb{E}\left[W \cdot \left\|  \hat I',I' \right\|_1 \right]
\end{equation}
where $\mathbb{E}(\cdot)$ is the expectation. Then, the confidence weight can reduce the impact of the mismatching area while maximizing the error in other regions.

We utilize the original optimization \cite{kerbl3Dgaussians} in Eq.~\ref{eq:original_optimization} for input views and confidence-guided optimization for pseudo views. The total loss function is:
\begin{equation}
    \mathcal{L} = \mathcal{L}_{ori} + \mathcal{L}_{con}
\end{equation}


\section{Experiment}
\subsection{RoadSight Dataset}
Existing autonomous driving datasets mainly rely on vehicle-mounted sensors, 
while neglecting the potential of infrastructure-based perception. Although \cite{v2x-seq} and \cite{dair-v2x} represent an infrastructure-vehicle cooperative dataset for improving 3D object detection, they cannot provide crossing views for scene reconstruction. Therefore, we introduce RoadSight, a multi-modality, multi-view dataset from real scenarios for I2V view transformation. This section describes the specifications of infrastructure sensors and how we set up these sensors.

\begin{figure}[t]
    \centering
    \includegraphics[width=0.8\linewidth]{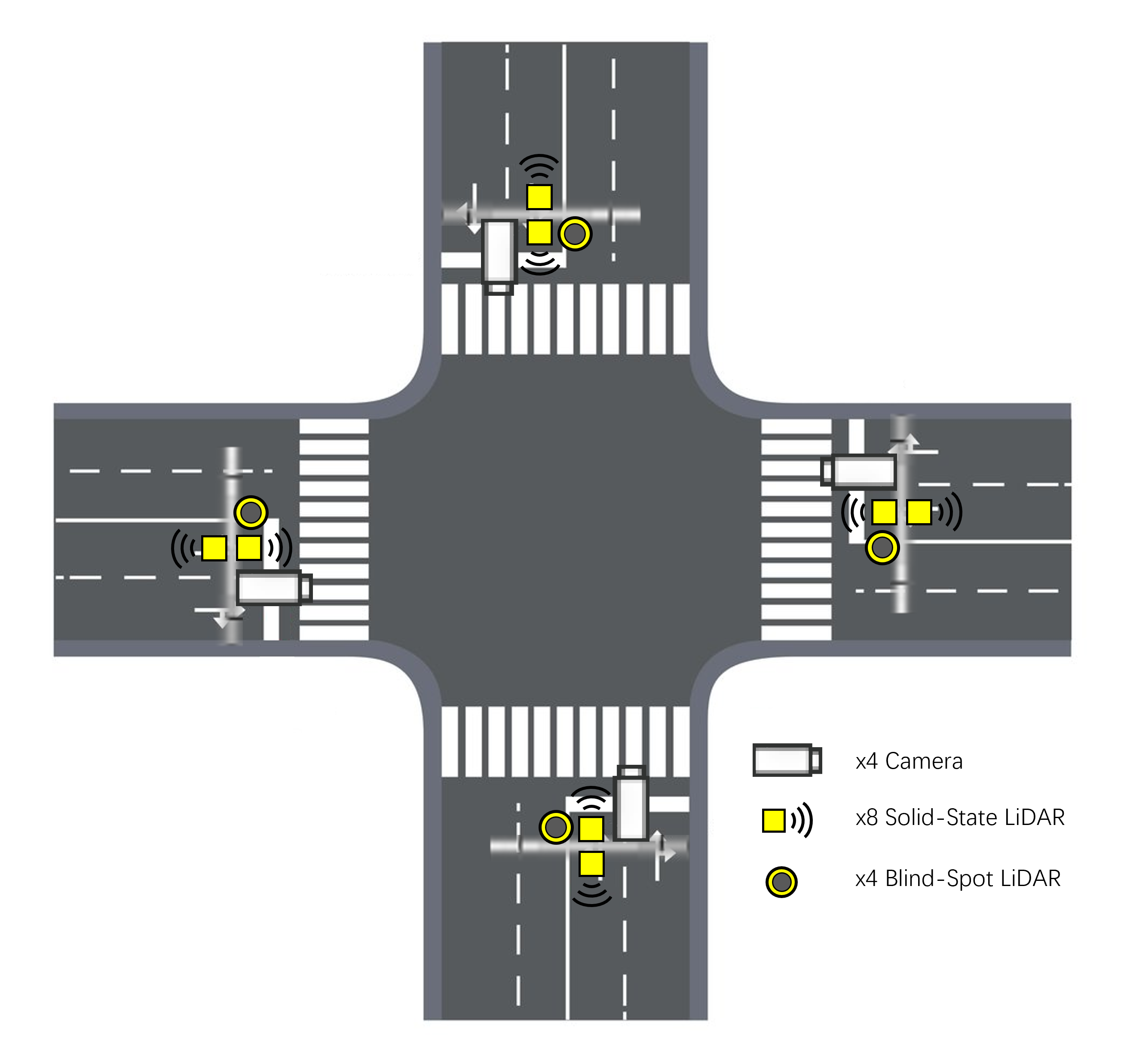}
    \caption{The layout of sensors. All sensors are attached to the traffic lights.}
    \label{fig:dataset}
\end{figure}

\textbf{Sensor Specification.} The data collection uses solid-state LiDAR, blind-spot LiDAR, and high-resolution cameras. The details of sensors are listed in Tab. \ref{tab:LiDAR_specification}. All sensors are hardware-synchronized via GPS-PPS signals, ensuring temporal alignment. Fig. \ref{fig:dataset} posts the specified layout. For all scenarios, sensors are attached to the traffic lights. The camera lens angles are roughly tilted down about 3 degrees and manually adjusted to align with the center of the road. Besides, an online extrinsic calibration module is developed to update the sensor poses and maintain the accuracy of data collection.




\textbf{Collection and Statistics.} RoadSight covers recordings from 4 urban intersections in Suzhou Autonomous Driving Demonstration Area, 
spanning diverse conditions:
\begin{itemize}
  \item \textbf{Traffic Density}: Peak hours (40\%), Off-peak (60\%);
  \item \textbf{Illumination}: Daytime (70\%), Night (30\%).
\end{itemize}
After raw data collection, we manually select 50 representative scenes, each lasting 20 seconds. The videos are sampled at 10Hz and synchronized with LiDAR scans.

\textbf{Privacy Protection.} RoadSight prioritizes ethical data collection and privacy protection. The whole dataset is collected from public roads with government authorization. Besides, we employ professional labeling tools to anonymize faces and license plates, ensuring the protection of personal identities.

\begin{table}[t]
    \centering
    \begin{tabular}{lccc}
\toprule
 &  LiDAR$^1$& LiDAR$^2$ & Camera\\
   \midrule
 Manufacturer& Innovusion& Bpearl&Hikivision\\
 Model& FALCON& RS&2CD7U8XJM\\
 Resolution&  -& -&1920$\times$1080\\
 Frequency& 10 Hz&10 Hz & 25 Hz\\
 HFOV& 120$^{\circ}$ & 360$^{\circ}$ & 89$^{\circ}$\\
 VFOV& 25$^{\circ}$ & 90$^{\circ}$ & 46.5$^{\circ}$\\
 Range& 500 m& 30 m &-\\ 
 Accuracy& $\pm$5 cm & $\pm$3 cm &-\\
\bottomrule
\end{tabular}
    \caption{Specification for Solid-State LiDAR (LiDAR$^1$), Blind-Spot LiDAR (LiDAR$^2$), and Camera. The HFOV and VFOV represent horizontal and vertical fields of view, respectively.}
    \label{tab:LiDAR_specification}
\end{table}

\subsection{Experiment Setup}
\textbf{Dataset.} We conduct the experiments on the RoadSight dataset, where 4 representative scenes are selected. Furthermore, to validate the robustness, we assess our method on 10 selected sequences with surrounding videos and LiDAR point clouds from Waymo Dataset~\cite{sun2020scalability}. For a fair comparison, we train the 3DGS-based methods using the first frame and 4DGS-based methods using the first 10 frames.  

\textbf{Metric.} Following \cite{zhao2024drivedreamer4d}, we adopt Novel Trajectory Agent IoU (NTA-IoU) and Novel Trajectory Lane IoU (NTL-IoU), which detect vehicles and road lanes in novel trajectory viewpoints and compare them with ground truth after projection. Additionally, we utilize the FID \cite{heusel2017gans} to assess the difference in feature distribution between the synthesized novel view and the original view.



\textbf{Implementation Details.} Our model is trained for 60,000 iterations with the Adam optimizer \cite{kingma2014adam}. We adopt Depth-Anything \cite{depth_anything_v2} as monocular depth estimation model. We initially warm up the optimization with Gaussian model~\cite{kerbl3Dgaussians} for 3,000 iterations and then generate pseudo views with a cascade strategy every 3,000 iterations for three cycles.

\begin{table*}[htbp]
\centering
\begin{tabular}{l|ccc|ccc}
\toprule
\multirow{2}{*}{} & \multicolumn{3}{c|}{RoadSight}& \multicolumn{3}{c}{Waymo \cite{sun2020scalability}} \\

 & NTA-IoU↑& NTL-IoU↑&  FID↓ & NTA-IoU↑& NTL-IoU↑& FID↓ \\
\midrule
DNGaussian~\cite{li2024dngaussian}
& 0.561& 50.02& 265.18& 0.491& 49.28& 89.91\\
SparseGS~\cite{xiong2023sparsegs}
& 0.554& 67.74& 224.62& 0.392& 49.27& 93.34\\
StreetGaussian~\cite{yan2024street}
& 0.552& 63.03& 231.84& 0.498& 50.19& 110.37\\
S\textsuperscript{3}Gaussian~\cite{huang2024textit}& 0.538& 59.13& 237.41& 0.384& 48.75& 130.43\\
I2V-GS (Ours)& \textbf{0.804}& \textbf{84.62}& \textbf{197.35}& \textbf{0.646}& \textbf{52.31}& \textbf{73.54}\\
\bottomrule
\end{tabular}
\caption{Comparison of NTA-Iou, NTL-Iou, and FID scores under novel view.}
\label{tab:comparison}
\end{table*}

\begin{figure*}[!t]
    \centering
    \includegraphics[width = \textwidth]{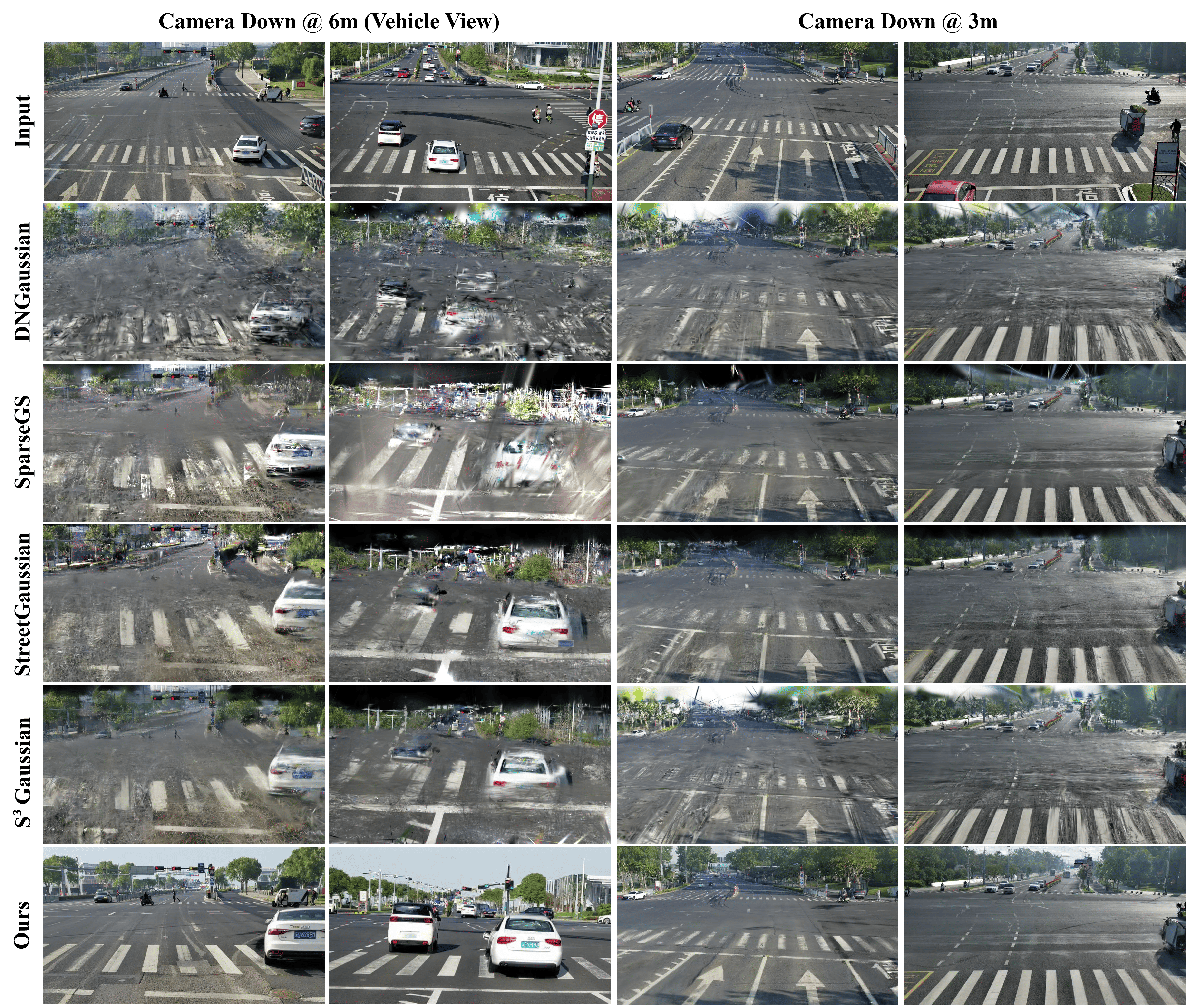}
    \caption{Quality comparison of vehicle view rendering with DNGaussian~\cite{li2024dngaussian}, SparseGS~\cite{xiong2023sparsegs}, StreetGaussian~\cite{yan2024street}, and S$^3$Gaussian~\cite{huang2024textit} on RoadSight dataset. Four input images are from the same intersection with the same timestamp.}
    \label{fig:result}
\end{figure*}

\begin{figure*}[t]
    \centering
    \includegraphics[width = \textwidth]{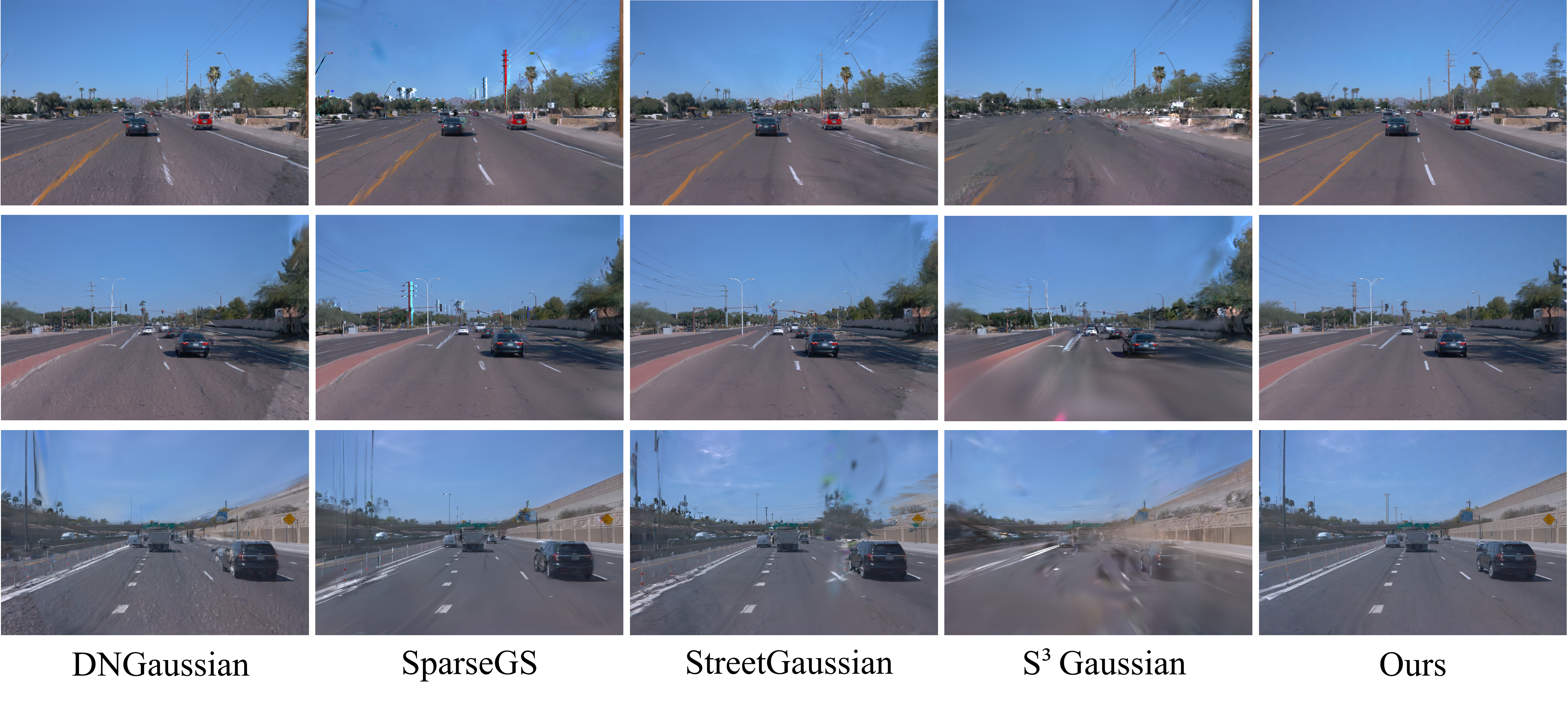}
    \vspace{-0.8 cm}
    \caption{Quality comparison of novel trajectory rendering with DNGaussian~\cite{li2024dngaussian}, SparseGS~\cite{xiong2023sparsegs}, StreetGaussian~\cite{yan2024street}, and S$^3$Gaussian~\cite{huang2024textit} on Waymo \cite{sun2020scalability} dataset. }
    \label{fig:result_waymo}
\end{figure*}

\begin{figure*}[t]
    \centering
    \includegraphics[width = \textwidth]{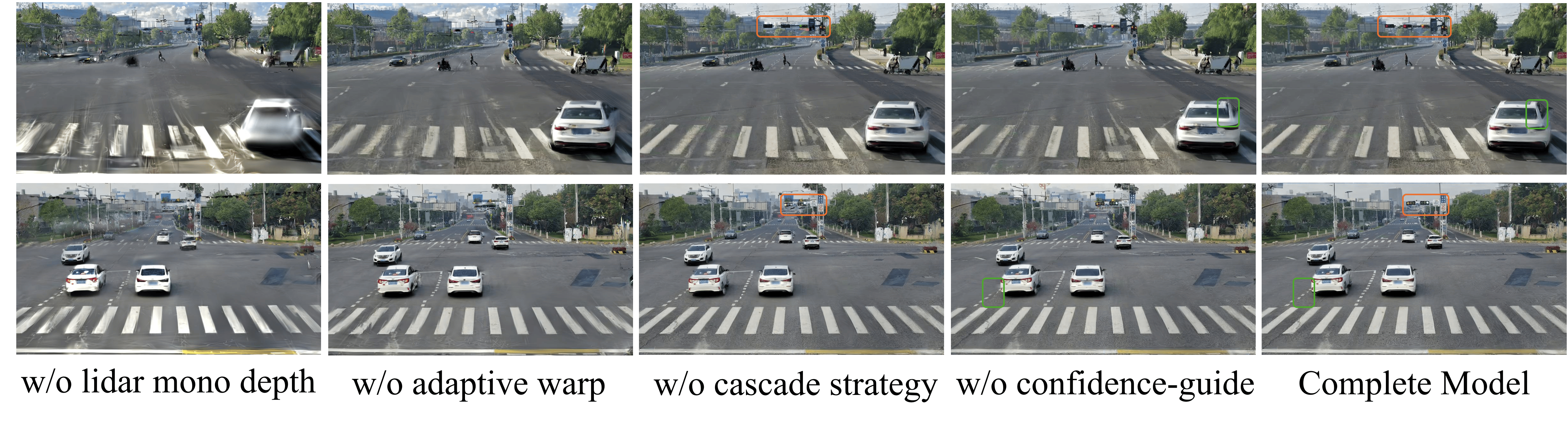}
    \vspace{-0.8 cm}
    \caption{Ablation study on proposed methods. Original monocular depth is adopted in `w/o LiDAR mono depth' and fixed depth warp is employed in `w/o adaptive warp'.}
    \label{fig:ablation}
\end{figure*}

\subsection{Comparison with State-of-the-art}
\textbf{Results on RoadSight Dataset.} We synthesize novel vehicle views from the captured images. Tab. \ref{tab:comparison} demonstrates our improvements on NTA-Iou, NTL-Iou, and FID, outperforming StreerGaussian \cite{yan2024street} with a 45.7\% increase in NTA-Iou, 34.2\% increase in NTL-Iou, and 14.9\% in FID. These enhancements are visually demonstrated in Fig. \ref{fig:result}. Our approach renders high-quality foreground vehicles and background elements in vehicle views, resulting in more realistic driving environments. In contrast, baseline methods suffer from artifacts and blurry results due to the lack of capacity of their model under large view transformation.

\textbf{Results on Waymo Dataset \cite{sun2020scalability}.} To validate the robustness of our approach, we conduct experiments on the Waymo dataset \cite{sun2020scalability}, where the novel view is synthesized along novel trajectories shifting from the recorded trajectories. The results are reported in Tab. \ref{tab:comparison}. Compared to StreetGaussian \cite{yan2024street}, our approach achieves 29.7\% NTA-Iou improvement, 4.2\% NTL-Iou improvement, and 33.3\% reduction FID. These achievements can be observed in Fig. \ref{fig:result_waymo}, where our approach demonstrates robust rendering quality under trajectory shifting.

\begin{table}[t]
\centering
\begin{tabular}{l|cc}
\toprule
 & NTA-IoU↑ & NTL-IoU↑ \\
\midrule
w/o LiDAR mono depth& 0.513& 43.86\\
w/o adaptive warp& 0.682& 74.41\\
w/o cascade strategy& 0.786& 83.49\\
 w/o confidence-guide& 0.783& 81.57\\
 Complete model& \textbf{0.804}& \textbf{84.62}\\
 \bottomrule
\end{tabular}
\caption{Ablation studies on proposed methods.}
\label{tab:ablation}
\end{table}

\subsection{Ablation Study}
To verify the effectiveness of the proposed method, we isolate each of these modules separately while keeping the other modules unchanged, evaluating the metrics and illustrating the visualization results. As shown in Tab. \ref{tab:ablation}, the performance decreases when replacing any of the modules.

\textbf{LiDAR Monocular Depth.} We adopt the original monocular depth and employ Pearson correlation loss in optimization to evaluate the effectiveness. From Fig. \ref{fig:ablation} `w/o LiDAR mono depth', it can be observed that monocular depth with Pearson loss fails to constrain the position of Gaussians, causing blurring and noise. 

\textbf{Adaptive Depth Warp.} We employ fixed depth warp in Fig. \ref{fig:ablation} `w/o adaptive warp'. Although the quality of the bottom image is acceptable, there are artifacts in the top image. This indicates that adaptive depth warp is robust to scenario changes to generate diverse training views.

\textbf{Cascade Strategy.} We separately inpaint holes for each warped image in Fig. \ref{fig:ablation} `w/o cascaded strategy'. The traffic light in the top image is significantly different from the input, where the red light is changed to a green light, while that in the bottom image is distorted. 

\textbf{Confidence-Guided Optimisation.} The confidence-guided optimisation is removed in Fig. \ref{fig:ablation} `w/o confidence-guide'. It is evident that confidence-guided optimization can avoid the inaccuracy of the diffusion model and improve the render quality.

\section{Conclusion}
In this paper, we present I2V-GS, the first framework for generating autonomous driving datasets with I2V view transformation. To address challenges caused by the sparse view input and large view transformation, we first adopt LiDAR to calibrate monocular depth to provide an accurate depth. Then these depths are utilized to carry the adaptive depth warp to generate dense training views. The cascade strategy is introduced to inpaint holes in warped images iteratively to ensure the consistency of inpaint content across views, which also enables the depth warp to generate a wider range of views. The cross-view information is employed to guide the optimization to ensure the reliability of pseudo views. Extensive experiments demonstrate that our approach significantly improves the view synthesis quality from the vehicle view. These results highlight the possibility of leveraging I2V-GS to generate training data for end-to-end autonomous driving systems.
{
    \small
    \bibliographystyle{ieeenat_fullname}
    \bibliography{main}
}


\end{document}